\documentclass[11pt, a4paper]{huawei}
\usepackage[numbers,sort&compress]{natbib}
\usepackage{dblfloatfix}
\usepackage{caption}
\usepackage{subcaption}
\usepackage{xspace}
\usepackage{pifont}
\usepackage{multirow}
\usepackage{tcolorbox}
\usepackage{hyperref}

\usepackage{amsfonts}
\usepackage{amsmath}
\usepackage{amssymb}
\usepackage{cleveref}
\usepackage{makecell}

\usepackage{dsfont}
\usepackage[table]{xcolor}
\usepackage{tabularx}
\usepackage{booktabs}
\newcolumntype{Y}{>{\centering\arraybackslash}X}

\usepackage{graphicx}
\usepackage{tikz}
\usetikzlibrary{shapes,arrows,positioning,fit,backgrounds,calc,shapes.geometric,decorations.pathreplacing,patterns}
\usepackage{algorithm}
\usepackage{algpseudocode}
\usepackage{float}
\usepackage{nicefrac}
\usepackage{microtype}
\usepackage[utf8]{inputenc}
\usepackage[T1]{fontenc}

\newcommand{\gain}[1]{{\scriptsize\color{green!50!black}(#1)}}

\definecolor{claudecolor}{RGB}{204, 119, 34}
\definecolor{evolvecolor}{RGB}{76, 114, 176}
\definecolor{datacolor}{RGB}{85, 168, 104}
\definecolor{evalcolor}{RGB}{196, 78, 82}
\definecolor{taskcolor}{RGB}{129, 114, 178}
\definecolor{skillcolor}{RGB}{218, 165, 32}
\definecolor{policycolor}{RGB}{70, 130, 180}

\title{EE-MCP: Self-Evolving MCP-GUI Agents via Automated Environment Generation and Experience Learning}

\author[1]{Tiantian He}
\author[1]{Yihang Chen}
\author[1]{Keyue Jiang}
\author[2]{Ka Yiu Lee}
\author[3]{Kaiwen Zhou}
\author[4$\dagger$]{Kun Shao}
\author[5$\dagger$]{Shuai Wang}

\affil[1]{University College London}
\affil[2]{Huawei Noah's Ark Lab}
\affil[3]{The Chinese University of Hong Kong}
\affil[4]{Independent Researcher}
\affil[5]{Peking University}



\begin{abstract}
Computer-use agents that combine GUI interaction with structured API calls via the Model Context Protocol (MCP) show promise for automating software tasks. However, existing approaches lack a principled understanding of how agents should balance these two modalities and how to enable iterative self-improvement across diverse applications. We formulate MCP-GUI interplay as a \textbf{unified hybrid policy learning problem} where the agent learns \textit{when} each modality provides complementary advantages, and show that distillation and experience augmentation target fundamentally different failure modes---requiring application-aware mechanism selection. Built on this formulation, we propose a \textbf{self-evolving framework} with a fully automatic pipeline that orchestrates \textit{automatic environment generation and validation}, trajectory collection, gap-driven task synthesis, and quality-filtered training---all without manual intervention. A key innovation is our \textbf{experience bank}, which accumulates LLM-learned rules from trajectory comparison, enabling inference-time improvement without fine-tuning. Systematic \textbf{cross-application analysis} across three desktop applications reveals that the optimal strategy depends on MCP--GUI composition: distillation achieves 77.8\% pass rate on MCP-dominant tasks (+17.8pp), while the experience bank excels on GUI-intensive tasks (+10.0pp).
\end{abstract}

\begin{document}

\maketitle

\begingroup
\renewcommand\thefootnote{$^{\dagger}$}%
\footnotetext{Corresponding authors: Shuai Wang (wangshuai\_2016@pku.edu.cn), Kun Shao (shaokun1991@gmail.com).}%
\endgroup

% ============================================
% 1. INTRODUCTION
% ============================================
\section{Introduction}
\label{sec:intro}

Computer-use agents (CUAs) powered by large language models have emerged as a practical paradigm for automating complex software tasks. These agents interact with applications through multiple modalities---visual perception of graphical user interfaces, structured API calls via tool protocols, and keyboard/mouse actions. Such hybrid interaction patterns are becoming increasingly common across diverse domains, from web automation to desktop productivity tools. The challenge lies not only in mastering individual modalities, but in learning \textit{when} to leverage each for complementary advantages.

The recent introduction of Model Context Protocol (MCP) \cite{anthropic2024mcp} exemplifies this trend, enabling agents to combine structured API calls with traditional GUI manipulation. This raises a fundamental question:
\begin{center}
\textit{How should agents learn to balance MCP tool calls and GUI actions, and what mechanisms enable effective self-improvement across diverse applications?}
\end{center}
Recent benchmarks such as MCPWorld \cite{yan2025mcpworld} and OSWorld-MCP \cite{jia2025osworldmcp} highlighted the importance of this question. MCPWorld demonstrates that hybrid agents combining MCP and GUI outperform both GUI-only and MCP-only approaches. OSWorld-MCP further shows that MCP tools can significantly improve agent accuracy, while also revealing that tool invocation rates remain low even for state-of-the-art models. These findings underscore the need for methods that help agents learn \textit{when} and \textit{how} to use each modality effectively.

Existing approaches to training computer-use agents rely on either one-shot supervised fine-tuning (SFT) from expert demonstrations or online reinforcement learning (RL) with environment rewards. While SFT enables basic skill acquisition and RL permits iterative improvement, both share key limitations: (1) they treat all training samples or rewards equally regardless of \textit{which} weaknesses most need correction, (2) neither diagnoses systematic failure patterns to generate targeted training data, and (3) they do not reveal how evolution mechanisms interact with application-specific MCP--GUI task compositions.

We propose a \textbf{self-evolving policy learning} framework (Figure~\ref{fig:overview}) that views agent learning through the lens of \textit{iterative policy refinement} for MCP-GUI synergetic collaboration. Rather than treating MCP tools and GUI actions as independent modalities, we formulate their interplay as a \textbf{unified hybrid policy learning problem} where the agent must learn \textit{when} each modality provides complementary advantages. The framework operates as a fully automatic closed-loop system: multi-dimensional performance profiling identifies systematic weaknesses, driving targeted task and environment generation; an experience bank accumulates actionable rules from LLM-based trajectory comparison for inference-time improvement; and the pipeline autonomously orchestrates environment setup, trajectory collection, and quality-filtered training---all without manual intervention.

A central finding is that the effectiveness of evolution mechanisms depends critically on the \textbf{MCP--GUI task composition} of each application. Chrome benefits most from distillation (+17.8pp), while GUI-intensive VS Code benefits primarily from the experience bank (+10pp). This requires \textit{application-aware} mechanism selection---a principle that generalizes to any hybrid agent training system.

\paragraph{Contributions.} We make the following contributions:

\begin{itemize}
    \item A \textbf{hybrid MCP-GUI policy learning formulation} that models the interplay between structured API calls and visual GUI actions as a unified sequential decision-making problem, where the agent learns \textit{when} to use each modality. We show that distillation and experience augmentation are complementary rather than interchangeable: distillation excels for MCP-dominant tasks while experience banks are more effective for GUI-intensive tasks, requiring application-aware mechanism selection.

    \item A \textbf{self-evolving framework} with a fully automatic pipeline that orchestrates expert trajectory collection, environment generation and validation, experience bank construction, LLM-judge evaluation, multi-dimensional gap analysis, and adaptive task generation---all without manual intervention.

    \item An \textbf{experience bank mechanism} that accumulates concise, actionable rules from LLM-based trajectory comparison, enabling inference-time improvement without additional fine-tuning. The bank is organized by skill category with capacity limits and filtered by application type to avoid cross-app contamination.

    \item Systematic \textbf{cross-application analysis} across Chrome, VS Code, and LibreOffice Calc revealing that the optimal evolution strategy depends on each application's MCP--GUI task composition, providing practical guidance for practitioners deploying hybrid agents.
\end{itemize}

% ============================================
% 2. RELATED WORK
% ============================================
\section{Related Work}
\label{sec:related}

\paragraph{Computer Use Agents and Benchmarks.}
The development of computer use agents has been driven by benchmarks that evaluate GUI interaction capabilities. OSWorld \cite{xie2024osworld} provides a comprehensive environment for testing agents on real desktop applications. WebArena \cite{zhou2024webarena} and VisualWebArena \cite{koh2024visualwebarena} focus on web-based tasks. Mind2Web \cite{deng2023mind2web} provides large-scale web interaction data.

Most relevant to our work, MCPWorld \cite{yan2025mcpworld} and OSWorld-MCP \cite{jia2025osworldmcp} introduce benchmarks specifically designed to evaluate hybrid MCP-GUI agents. MCPWorld presents tasks across multiple desktop applications with curated MCP tools, demonstrating that hybrid approaches outperform single-modality agents. OSWorld-MCP provides tool-beneficial tasks with automated tool generation pipelines, revealing that tool invocation rates remain low even for state-of-the-art models. While these benchmarks characterize \textit{what} agents can do, our work focuses on \textit{how} agents can learn to balance modalities through self-evolution.

\paragraph{Learning from Demonstrations.}
Agent training from demonstrations has been explored extensively. AgentTrek \cite{xu2024agenttrek} uses web-based trajectory collection. DigiRL \cite{bai2024digirl} combines offline and online learning for device control. Agent Workflow Memory \cite{wang2024agent} extracts reusable procedures from demonstrations. Our approach differs by using Claude as an expert demonstrator and incorporating iterative refinement based on multi-dimensional performance feedback.

\paragraph{Self-Improvement and Iterative Learning.}
Self-improvement methods for LLMs have shown promise in various domains. STaR \cite{zelikman2022star} bootstraps reasoning by iteratively fine-tuning on self-generated rationales, using rationalization to recover from failure cases. Self-Instruct \cite{wang2023selfinstruct} generates training data from model outputs. ReST \cite{gulcehre2023rest} applies reinforcement learning from self-play. Our framework adapts these ideas to computer use agents, using structured performance profiles to drive targeted data generation. Our self-improvement variant (Section~\ref{sec:self_improvement}) follows the rejection sampling principle: training only on the model's own successful outputs to avoid distribution shift.

% ============================================
% 3. PROBLEM FORMULATION
% ============================================
\section{Problem Formulation}
\label{sec:formulation}

\subsection{Computer Use as Sequential Decision Making}

We formulate computer use as a Markov Decision Process (MDP) where an agent interacts with applications through both MCP tools and GUI actions. The central challenge is learning a \textbf{hybrid policy} that selects the optimal modality at each step---a decision that depends on the application, task type, and current state.

\paragraph{State Space.}
Each state $s \in \mathcal{S}$ comprises the current screenshot $s_{\text{visual}} \in \mathbb{R}^{H \times W \times 3}$, available MCP tool descriptions $s_{\text{mcp}}$, and action history $s_{\text{history}}$.

\paragraph{Action Space.}
The agent selects from two modalities at each step:

\textit{MCP Actions} $\mathcal{A}_{\text{MCP}}$: Structured API calls following the Model Context Protocol format:
\begin{equation}
a_{\text{mcp}} = \langle \texttt{tool\_name}, \texttt{arguments} \rangle
\end{equation}

\textit{GUI Actions} $\mathcal{A}_{\text{GUI}}$: Visual interface interactions including click, type, scroll, and keyboard shortcuts:
\begin{equation}
a_{\text{gui}} = \langle \texttt{action\_type}, \texttt{coordinates}, \texttt{parameters} \rangle
\end{equation}

\paragraph{Task Specification and Evaluation.}
A task $\tau = \langle g, s_0, d, \mathbf{c}, \phi_{\text{judge}} \rangle$ consists of a natural language goal $g$, initial state $s_0$, difficulty level $d \in \{\text{easy}, \text{medium}, \text{hard}\}$, required skill categories $\mathbf{c} \subseteq \mathcal{C}$, and an LLM-judge evaluation function $\phi_{\text{judge}}: \xi \rightarrow [0,1]$.

\subsection{Skill Categories}
\label{sec:skills}

We define six \textbf{application-agnostic} skill categories $\mathcal{C}$ that characterize different aspects of computer use:
\textbf{Data Retrieval} ($c_{\text{retrieve}}$)---reading files, querying databases;
\textbf{Data Manipulation} ($c_{\text{manipulate}}$)---writing, editing, creating data;
\textbf{Search \& Query} ($c_{\text{search}}$)---finding patterns, filtering results;
\textbf{Execution \& Automation} ($c_{\text{exec}}$)---running scripts, triggering actions;
\textbf{Navigation \& Browsing} ($c_{\text{nav}}$)---moving between views and UI elements;
\textbf{Configuration \& Settings} ($c_{\text{config}}$)---modifying preferences and parameters.
These categories apply across all applications in our benchmark (Table~\ref{tab:apps}); detailed definitions with cross-application examples are provided in Appendix~\ref{app:skills}.

\subsection{Two-Stage Policy Iteration Framework}

We formulate agent learning as a two-stage iterative framework rather than one-shot imitation. Let $\pi_k$ denote the agent policy at iteration $k$. Each iteration consists of two stages:

\paragraph{Stage 1: Policy Evaluation via Multi-Dimensional Profiling.}
The current policy $\pi_{k-1}$ is evaluated on a task set $\mathcal{G}$ using an LLM judge $\phi_{\text{judge}}$, producing a performance profile $\mathcal{P}_k$ that tracks capabilities across modality balance, difficulty levels, and skill categories (Section~\ref{sec:profiling}). This profile serves as the ``value function'' that identifies where the policy is weak.

\paragraph{Stage 2: Policy Improvement.}
Based on the profile $\mathcal{P}_k$, gap analysis identifies weaknesses and generates new tasks $\mathcal{G}_k$. The policy is then improved through trajectory distillation and experience accumulation (detailed in Section~\ref{sec:method}). The training objective is:
\begin{equation}
\pi_{k+1} = \arg\max_\pi \mathbb{E}_{\xi \sim \mathcal{D}_k}[\log \pi(a^* | s, g)]
\label{eq:policy_update}
\end{equation}
where $\mathcal{D}_k$ is the accumulated training distribution and $a^*$ denotes demonstrated actions---from the expert in distillation, or from the student's own successes in self-improvement (Section~\ref{sec:self_improvement}). The training distribution evolves across iterations:
\begin{equation}
\mathcal{D}_{k+1} = \mathcal{D}_k \cup \mathcal{D}_{\text{new}}(\mathcal{G}_k, \mathcal{P}_k)
\end{equation}
where $\mathcal{D}_{\text{new}}$ contains new expert trajectories collected on tasks $\mathcal{G}_k$ generated based on performance profile $\mathcal{P}_k$.

% ============================================
% 4. METHOD
% ============================================
\section{Method: Self-Evolving Policy Learning}
\label{sec:method}

% ============================================
% FIGURE 1: SELF-EVOLUTION FRAMEWORK OVERVIEW
% ============================================

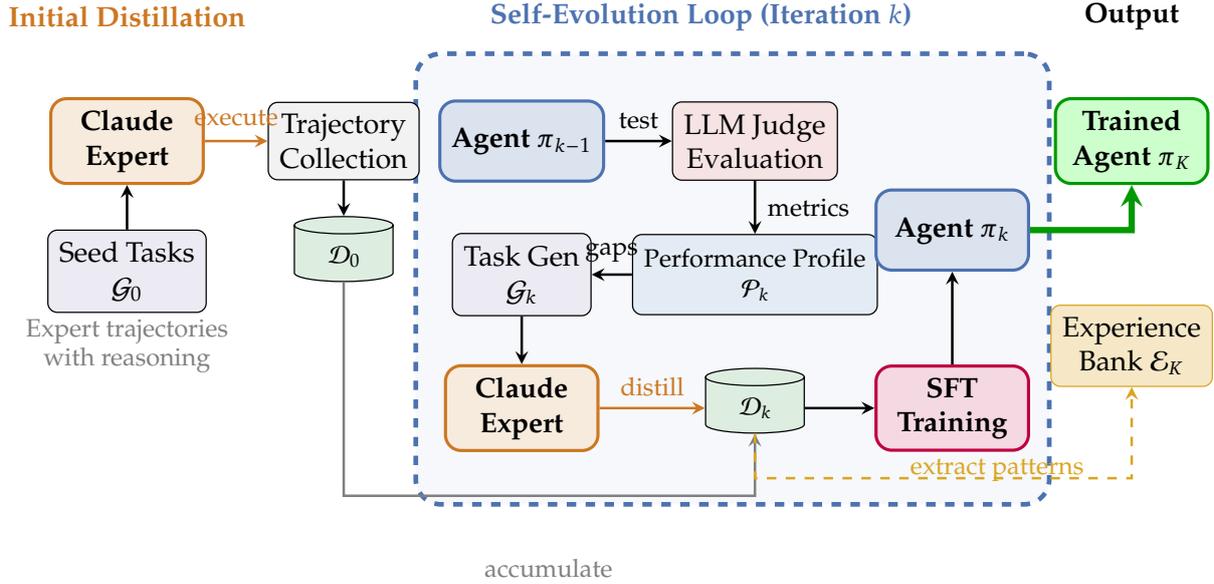
\begin{figure}[t]
\centering
\resizebox{\columnwidth}{!}{%
\begin{tikzpicture}[
    node distance=0.5cm and 0.6cm,
    box/.style={rectangle, draw, rounded corners=3pt, minimum height=0.8cm, minimum width=1.5cm, align=center, font=\footnotesize},
    mainbox/.style={rectangle, draw, rounded corners=5pt, minimum height=0.9cm, minimum width=1.7cm, align=center, font=\footnotesize\bfseries, line width=1pt},
    databox/.style={cylinder, draw, shape border rotate=90, aspect=0.3, minimum height=0.6cm, minimum width=1.1cm, font=\scriptsize},
    profilebox/.style={rectangle, draw, rounded corners=3pt, fill=policycolor!15, minimum height=0.9cm, minimum width=1.7cm, align=center, font=\scriptsize},
    arrow/.style={->, >=stealth, thick},
    bigarrow/.style={->, >=stealth, line width=2pt},
]

% ==========================================
% LEFT SECTION: Initial Distillation (Iter 0)
% ==========================================
\node[font=\footnotesize\bfseries, color=claudecolor] (title0) at (-4.2, 3.2) {Initial Distillation};

% Claude expert
\node[mainbox, fill=claudecolor!15, draw=claudecolor] (claude) at (-4.2, 1.8) {Claude\\Expert};

% Initial tasks
\node[box, fill=taskcolor!15, below=0.5cm of claude] (tasks0) {Seed Tasks\\$\mathcal{G}_0$};

% Trajectory collection
\node[box, fill=gray!10, right=0.7cm of claude] (collect0) {Trajectory\\Collection};

% Initial dataset
\node[databox, fill=datacolor!20, below=0.4cm of collect0] (D0) {$\mathcal{D}_0$};

% Arrows for initial distillation
\draw[arrow] (tasks0.north) -- (claude.south);
\draw[arrow, color=claudecolor] (claude) -- node[above, font=\scriptsize, yshift=1pt] {execute} (collect0);
\draw[arrow] (collect0) -- (D0);

% Annotation
\node[font=\scriptsize, text=gray, align=center] at (-4.2, -0.5) {Expert trajectories\\with reasoning};

% ==========================================
% CENTER SECTION: Self-Evolution Loop
% ==========================================
\node[font=\footnotesize\bfseries, color=evolvecolor] (titleloop) at (2.2, 3.2) {Self-Evolution Loop (Iteration $k$)};

% Evolution iteration box
\begin{scope}[on background layer]
\node[draw=evolvecolor, dashed, rounded corners=8pt, line width=1.5pt,
      inner sep=8pt, fill=evolvecolor!5,
      fit={(-0.7,-2.0) (5.8,2.5)}] (loopbox) {};
\end{scope}

% Row 1: Agent k-1 -> Eval
\node[mainbox, fill=evolvecolor!20, draw=evolvecolor] (agent) at (0.2, 1.8) {Agent $\pi_{k-1}$};
\node[box, fill=evalcolor!15] (eval) at (2.8, 1.8) {LLM Judge\\Evaluation};

% Row 2: Task Gen -> Profile -> Agent k
\node[box, fill=taskcolor!15] (taskgen) at (0.2, 0.3) {Task Gen\\$\mathcal{G}_k$};
\node[profilebox] (profile) at (2.8, 0.3) {Performance Profile\\$\mathcal{P}_k$};
\node[mainbox, fill=evolvecolor!20, draw=evolvecolor] (agentnew) at (5, 0.8) {Agent $\pi_k$};

% Row 3: Claude -> Dk -> SFT
\node[mainbox, fill=claudecolor!15, draw=claudecolor] (claudek) at (0.2, -1.2) {Claude\\Expert};
\node[databox, fill=datacolor!20] (Dk) at (2.8, -1.2) {$\mathcal{D}_k$};
\node[mainbox, fill=purple!15, draw=purple] (sft) at (5, -1.2) {SFT\\Training};

% Evolution loop arrows - all horizontal/vertical
% Agent k-1 -> Eval (right)
\draw[arrow] (agent.east) -- node[above, font=\scriptsize] {test} (eval.west);

% Eval -> Profile (down)
\draw[arrow] (eval.south) -- node[right, font=\scriptsize] {metrics} (profile.north);

% Profile -> Task Gen (left)
\draw[arrow] (profile.west) -- node[above, font=\scriptsize] {gaps} (taskgen.east);

% Task Gen -> Claude (down)
\draw[arrow] (taskgen.south) -- (claudek.north);

% Claude -> Dk (right)
\draw[arrow, color=claudecolor] (claudek.east) -- node[above, font=\scriptsize] {distill} (Dk.west);

% Dk -> SFT (right)
\draw[arrow] (Dk.east) -- (sft.west);

% SFT -> Agent k (up)
\draw[arrow] (sft.north) -- (agentnew.south);

% ==========================================
% Data accumulation from D0 to Dk
% D0 -> down -> right -> up -> Dk (using coordinates)
% ==========================================
\draw[arrow, gray] (D0.south) -- ++(0,-2.4) -| (Dk.south);
\node[font=\scriptsize, text=gray] at (0.5, -3.0) {accumulate};

% ==========================================
% RIGHT SECTION: Output
% ==========================================
\node[font=\footnotesize\bfseries] (titleout) at (7, 3.2) {Output};

% Final model
\node[mainbox, fill=green!20, draw=green!60!black] (final) at (7, 1.8) {Trained\\Agent $\pi_K$};

% Experience Bank output
\node[box, fill=skillcolor!25, draw=skillcolor] (expbank) at (7, -0.5) {Experience\\Bank $\mathcal{E}_K$};

% Arrow from Agent k to final: right to x=7, then up to final.south
\draw[bigarrow, color=green!60!black] (agentnew.east) -- (7, 0.8) -- (final.south);

% Arrow from experience synthesis to experience bank output
\draw[arrow, dashed, color=skillcolor] (Dk.south) -- ++(0,-0.5) -| (expbank.south);
\node[font=\scriptsize, text=skillcolor] at (5.5, -1.9) {extract patterns};

\end{tikzpicture}%
}%
\caption{
\textbf{Self-Evolving Policy Learning Framework.}
\textbf{Left}: Expert trajectories are collected from Claude on seed tasks $\mathcal{G}_0$.
\textbf{Center}: The self-evolution loop---the agent is evaluated via LLM judge to produce a multi-dimensional performance profile $\mathcal{P}_k$ (tracking MCP/GUI ratio, difficulty, and skills), which drives targeted task and environment generation. New trajectories are accumulated with existing data for SFT training.
\textbf{Right}: Two complementary outputs: (1) a fine-tuned agent $\pi_K$ (effective for MCP-dominant tasks), and (2) an experience bank $\mathcal{E}_K$ that enables inference-time improvement (effective for GUI-intensive tasks).
}
\label{fig:overview}
\end{figure}

Our framework operates as an iterative improvement loop (Figure~\ref{fig:overview}), with each component addressing a distinct aspect of hybrid policy learning. The two-stage structure (evaluation via profiling, improvement via SFT) mirrors policy iteration but operates offline with expert demonstrations rather than online exploration.

\subsection{Expert Trajectory Collection}

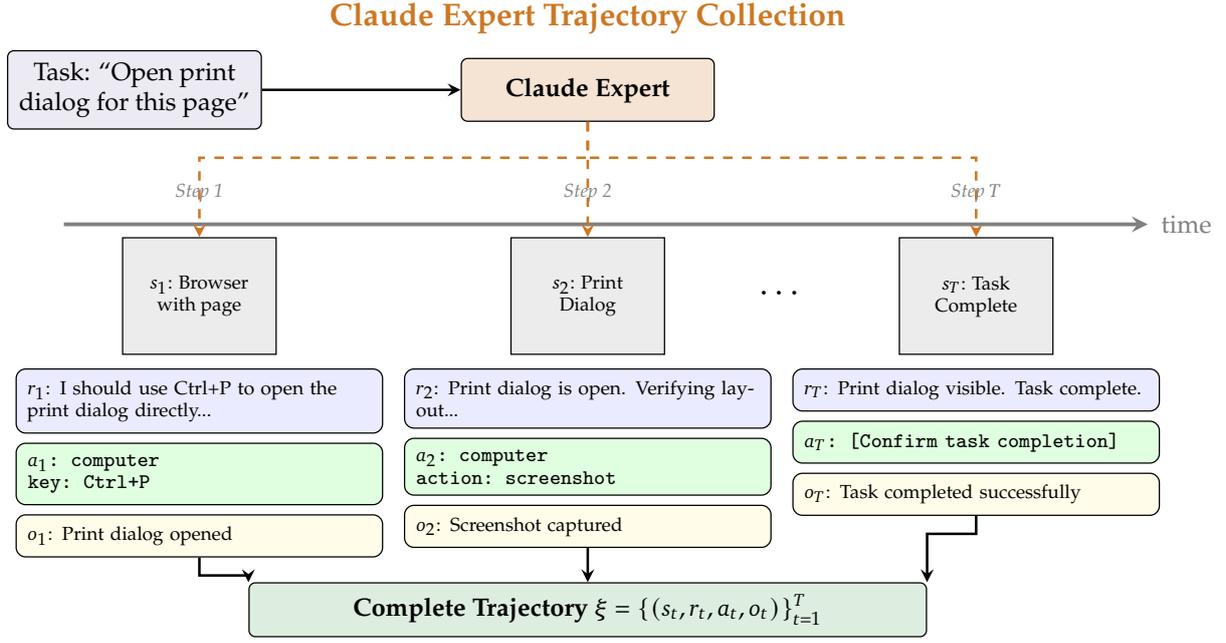
\begin{figure}[t]
\centering
\resizebox{\columnwidth}{!}{%
\begin{tikzpicture}[
    node distance=0.3cm,
    box/.style={rectangle, draw, rounded corners=2pt, minimum height=0.6cm, align=center, font=\scriptsize},
    screenshot/.style={rectangle, draw, fill=gray!15, minimum height=1.3cm, minimum width=1.7cm, align=center, font=\tiny},
    thought/.style={rectangle, draw, rounded corners=3pt, fill=blue!8, minimum height=0.45cm, align=left, font=\tiny, text width=3.8cm},
    action/.style={rectangle, draw, rounded corners=3pt, fill=green!12, minimum height=0.45cm, align=left, font=\tiny\ttfamily, text width=3.8cm},
    obs/.style={rectangle, draw, rounded corners=3pt, fill=yellow!10, minimum height=0.35cm, align=left, font=\tiny, text width=3.8cm},
    arrow/.style={->, >=stealth, thick},
    timeline/.style={->, >=stealth, very thick, color=gray},
]

\node[font=\small\bfseries, color=claudecolor] at (0, 3) {Claude Expert Trajectory Collection};
\node[box, fill=claudecolor!20, minimum width=2.8cm, minimum height=0.7cm] (claude) at (0, 2.2) {\textbf{Claude Expert}};
\node[box, fill=taskcolor!15, left=2.2cm of claude, minimum width=2.3cm] (task) {Task: ``Open print\\dialog for this page''};
\draw[arrow] (task) -- (claude);
\draw[timeline] (-5.8, 0.7) -- (6.2, 0.7) node[right, font=\scriptsize] {time};

\node[screenshot] (s1) at (-4.3, -0.1) {$s_1$: Browser\\with page};
\node[thought, below=0.15cm of s1] (r1) {$r_1$: I should use Ctrl+P to open the print dialog directly...};
\node[action, below=0.1cm of r1] (a1) {$a_1$: \texttt{computer}\\key: Ctrl+P};
\node[obs, below=0.1cm of a1] (o1) {$o_1$: Print dialog opened};

\node[screenshot] (s2) at (0, -0.1) {$s_2$: Print\\Dialog};
\node[thought, below=0.15cm of s2] (r2) {$r_2$: Print dialog is open. Verifying layout...};
\node[action, below=0.1cm of r2] (a2) {$a_2$: \texttt{computer}\\action: screenshot};
\node[obs, below=0.1cm of a2] (o2) {$o_2$: Screenshot captured};

\node[screenshot] (st) at (4.3, -0.1) {$s_T$: Task\\Complete};
\node[thought, below=0.15cm of st] (rt) {$r_T$: Print dialog visible. Task complete.};
\node[action, below=0.1cm of rt] (at) {$a_T$: [Confirm task completion]};
\node[obs, below=0.1cm of at] (ot) {$o_T$: Task completed successfully};

\node at (2.15, -0.1) {$\cdots$};
\draw[arrow, color=claudecolor, dashed] (claude.south) -- ++(0,-0.4) -| (s1.north);
\draw[arrow, color=claudecolor, dashed] (claude.south) -- ++(0,-0.4) -| (s2.north);
\draw[arrow, color=claudecolor, dashed] (claude.south) -- ++(0,-0.4) -| (st.north);

\node[box, fill=datacolor!20, minimum width=7.5cm, minimum height=0.6cm] (traj) at (0, -3.6) {\textbf{Complete Trajectory} $\xi = \{(s_t, r_t, a_t, o_t)\}_{t=1}^{T}$};
\draw[arrow] (o1.south) -- ++(0,-0.2) -| (traj.north west);
\draw[arrow] (o2.south) -- (traj.north);
\draw[arrow] (ot.south) -- ++(0,-0.2) -| (traj.north east);

\node[font=\tiny\itshape, color=gray] at (-4.3, 1.05) {Step 1};
\node[font=\tiny\itshape, color=gray] at (0, 1.05) {Step 2};
\node[font=\tiny\itshape, color=gray] at (4.3, 1.05) {Step $T$};
\end{tikzpicture}%
}%
\caption{
\textbf{Expert Trajectory Distillation.}
Claude executes tasks step-by-step, producing trajectories with screenshots~$s_t$, reasoning~$r_t$, actions~$a_t$ (MCP or GUI), and observations~$o_t$. These serve as training data for the student model.
}
\label{fig:distillation}
\end{figure}

We use Claude Sonnet 4 as the expert policy to collect trajectories that capture the complete decision-making process (Figure~\ref{fig:distillation}). Each trajectory $\xi = \{(s_t, r_t, a_t, o_t)\}_{t=1}^{T}$ records the screenshot $s_t$, reasoning $r_t$, action $a_t$ (MCP or GUI), and observation $o_t$. Tasks are attempted multiple times, retaining the highest-scoring trajectory. All trajectories are standardized to the model's native \texttt{<tool\_call>} format.

\subsection{Offline Policy Improvement}
\label{sec:policy_improvement}

This section describes the \textit{policy improvement} stage in our iterative refinement framework. Given the performance profile $\mathcal{P}_k$ from evaluation (Section~\ref{sec:profiling}), we improve the policy through two complementary mechanisms: (1) supervised fine-tuning on expert and self-generated trajectories (this section and Section~\ref{sec:self_improvement}), and (2) inference-time experience augmentation (Section~\ref{sec:self_improvement}). Here we focus on the trajectory distillation component.

\paragraph{Training Objective.}
Given accumulated trajectories $\mathcal{D}_k = \{\xi_i\}_{i=1}^{N_k}$, we minimize:
\begin{equation}
\mathcal{L}_{\text{SFT}}(\theta) = -\sum_{\xi \in \mathcal{D}_k} \sum_{t=1}^{|\xi|} \log p_\theta(r_t, a_t | s_{\leq t}, g)
\end{equation}

\paragraph{Data Accumulation Strategy.}
We accumulate trajectories across iterations to prevent catastrophic forgetting:
\begin{equation}
\mathcal{D}_{k} = \mathcal{D}_0 \cup \bigcup_{i=1}^{k-1} \mathcal{D}_{\text{new}}^{(i)}
\end{equation}

Critically, we reset to the base model $\theta_0$ at each iteration rather than continuing from $\pi_{k-1}$, preventing error accumulation. To further mitigate forgetting, we mix successful trajectories from previous iterations with newly collected data via memory replay.

\subsection{Policy Evaluation via Multi-Dimensional Profiling}
\label{sec:profiling}

This stage corresponds to the \textit{policy evaluation} step in our iterative refinement framework. Rather than computing a value function as in standard RL, we use an LLM judge to evaluate the current policy across multiple dimensions, identifying systematic weaknesses that guide the next improvement iteration.

% ============================================
% TABLE: PERFORMANCE PROFILE DIMENSIONS
% ============================================

\begin{table}[t]
\centering
\caption{Multi-dimensional performance profile $\mathcal{P}_k$ tracks agent capabilities across five dimensions. These metrics feed into gap analysis for targeted task generation.}
\label{fig:profile}
\small
\begin{tabular}{lp{9cm}}
\toprule
\textbf{Dimension} & \textbf{Metrics} \\
\midrule
1. Modality Balance & $\rho_{\text{mcp}}$, $\rho_{\text{gui}}$, $\rho_{\text{hybrid}}$ \\
2. Difficulty Level & $\sigma_{\text{easy}}$, $\sigma_{\text{medium}}$, $\sigma_{\text{hard}}$ \\
3. Skill Categories & $\sigma_{\text{retrieve}}$, $\sigma_{\text{manip}}$, $\sigma_{\text{search}}$, $\sigma_{\text{exec}}$, $\sigma_{\text{nav}}$, $\sigma_{\text{config}}$ \\
4. Format Quality & $\phi_{\text{format}}$, $\phi_{\text{parse}}$, $\phi_{\text{args}}$ \\
5. Efficiency & $\bar{n}_{\text{steps}}$, $\phi_{\text{complete}}$, $\phi_{\text{timeout}}$ \\
\midrule
\multicolumn{2}{l}{\textit{Output: Gap Analysis $\rightarrow$ Task Generation}} \\
\bottomrule
\end{tabular}
\end{table}

The performance profile $\mathcal{P}_k = \langle \boldsymbol{\rho}, \boldsymbol{\sigma}_d, \boldsymbol{\sigma}_c, \boldsymbol{\phi}, \boldsymbol{\eta} \rangle$ captures modality balance, difficulty-level success rates, per-skill success rates, format quality, and efficiency metrics (Table~\ref{fig:profile}).

\subsection{Profile-Guided Task and Environment Generation}

Based on the performance profile, we generate targeted tasks to address identified weaknesses.

\paragraph{Gap Analysis.}
We identify gaps by comparing profile metrics against target thresholds:
\begin{align}
\Delta_{\text{mcp}} &= \rho_{\text{target}} - \rho_{\text{mcp}} & \text{(modality gap)} \\
\Delta_d &= \gamma_d - \sigma_d & \text{(difficulty gap)} \\
\Delta_c &= \gamma_c - \sigma_c & \text{(skill gap)}
\end{align}
where $\rho_{\text{target}}$, $\gamma_d$, $\gamma_c$ are configurable thresholds. Tasks are generated weighted toward identified weaknesses using LLM-based synthesis. For example, at iteration~2 on Chrome the gap analysis identified the agent as GUI-heavy with underused tools (\texttt{bookmark\_page}, \texttt{delete\_browsing\_data}), generating targeted tasks to address these MCP gaps, with expert demonstrations collected for each. Generated task examples are in Appendix~\ref{app:additional_analysis}.

\paragraph{Automatic Environment Generation.}
A critical enabler of fully automatic evolution is \textbf{environment generation}: each generated task requires a valid application state (e.g., specific browser tabs, spreadsheet data, or open files) before evaluation can begin. Our pipeline automatically generates environment setup scripts from the task description using an LLM, then validates them by checking that the target application reaches the expected initial state. Failed environments are retried with error feedback. This eliminates the manual environment authoring bottleneck that limits scaling in existing benchmarks---each new task produced by gap analysis is immediately executable without human configuration. Details on the environment validation pipeline are in Appendix~\ref{app:pipeline_details}.

\subsection{Self-Improvement and Experience Bank}
\label{sec:self_improvement}

A key challenge with expert distillation is the \textbf{capability gap}: the expert (Claude) can successfully execute actions that the student model cannot reliably reproduce. Training on such trajectories introduces distribution shift and may cause the student to attempt actions beyond its capabilities.

We address this with a \textbf{self-improvement paradigm} that trains exclusively on the model's own successful trajectories:

\paragraph{Rejection Sampling.}
At each iteration, we run the current model $\pi_{k-1}$ on each task up to $N=3$ attempts, retain trajectories exceeding a success threshold $\tau_{\text{success}} = 0.5$, select the best trajectory per task by LLM-judge score, and fine-tune on these self-generated trajectories.

\paragraph{Experience Bank: Cross-Iteration Knowledge Transfer.}
A key innovation is the \textbf{experience bank}---a structured knowledge base that accumulates LLM-learned patterns across iterations. Unlike simple trajectory replay, the experience bank uses an LLM to \textit{compare} successful and failed trajectories, extracting transferable insights about what strategies work.

The experience bank is organized by \textbf{skill category} (Section~\ref{sec:skills}), with each category maintaining four types of knowledge: (1) \textit{task-specific strategies}---patterns that led to success (e.g., ``take a screenshot first to identify the current view''); (2) \textit{environment knowledge}---LLM-learned observations about UI layout and application behavior; (3) \textit{tool usage patterns}---effective MCP tool sequences for common operations; and (4) \textit{error recovery strategies}---approaches that helped recover from failures.

\paragraph{Experience Bank Construction.}
At the end of each iteration, we build the experience bank by: (1) collecting successful trajectories from LLMs; (2) grouping by skill category; (3) comparing successful vs.\ failed trajectories using an LLM judge; (4) synthesizing concise, transferable patterns; and (5) filtering by application type to prevent cross-app contamination. Formally, for each skill category $c \in \mathcal{C}$, we store $\mathcal{E}_c = \langle \mathbf{s}_c, \mathbf{e}_c, \mathbf{t}_c \rangle$ (strategies, environment knowledge, tool patterns). Given successful trajectory $\xi^+$ and failed trajectory $\xi^-$, we extract differentiating factors via $(\mathbf{s}, \mathbf{e}, \mathbf{t}) = \text{LLM}_{\text{extract}}(\xi^+, \xi^-, c)$. When the bank exceeds a per-category capacity limit, we merge via LLM summarization: $\mathcal{E}_c^{\text{new}} = \text{LLM}_{\text{merge}}(\mathcal{E}_c^{\text{old}}, \mathcal{E}_c^{\text{fresh}})$.

\paragraph{Experience-Enhanced Prompts.}
At inference time, prompts are augmented with relevant experience:
\begin{equation}
\text{prompt}_k = \text{prompt}_{\text{base}} \oplus \text{Experience}(\mathcal{E}_{k-1}, c_{\text{task}})
\end{equation}
where $\mathcal{E}_{k-1}$ is the experience bank and $c_{\text{task}}$ is the skill category of the current task. This enables \textbf{inference-time improvement}---agents can benefit from accumulated experience without requiring additional fine-tuning.

\paragraph{Evolution Modes.}
We evaluate two modes: \textbf{Distillation + Experience} (SFT with experience-enhanced prompts) and \textbf{Experience-Only} (experience bank augmentation without fine-tuning, suitable when compute is limited).

\subsection{Self-Evolution Algorithm}
\label{sec:automatic_pipeline}

Algorithm~\ref{alg:evolution} summarizes the complete self-evolution pipeline. Infrastructure details including environment validation and distributed execution are in Appendix~\ref{app:pipeline_details}.

\begin{algorithm}[t]
\caption{Self-Evolution Pipeline}
\label{alg:evolution}
\begin{algorithmic}[1]
\State \textbf{Input:} Seed tasks $\mathcal{G}_0$, base model $\theta_0$, iterations $K$
\State \textbf{Output:} Trained agent $\pi_K$, experience bank $\mathcal{E}_K$
\State $\mathcal{D}_0 \gets \text{CollectExpert}(\text{Claude}, \mathcal{G}_0)$ \Comment{Phase 1}
\State $\pi_0 \gets \text{SFT}(\theta_0, \mathcal{D}_0)$ \Comment{Phase 2 (optional)}
\State $\mathcal{E}_0 \gets \text{BuildBank}(\mathcal{D}_0)$ \Comment{Initial bank}
\For{$k = 1$ to $K$}
    \State $\mathcal{P}_k \gets \text{Evaluate}(\pi_{k-1}, \mathcal{G}_0, \mathcal{E}_{k-1})$ \Comment{Policy Evaluation}
    \State $\mathcal{G}_k \gets \text{GenerateTasks}(\mathcal{P}_k)$ \Comment{Gap-driven}
    \State $\mathcal{D}_k^{\text{new}} \gets \text{CollectExpert}(\text{Claude}, \mathcal{G}_k)$
    \State $\mathcal{D}_k \gets \mathcal{D}_{k-1} \cup \mathcal{D}_k^{\text{new}}$
    \State $\pi_k \gets \text{SFT}(\theta_0, \mathcal{D}_k)$ \Comment{Policy Improvement}
    \State $\mathcal{E}_k \gets \text{BuildBank}(\mathcal{D}_k, \mathcal{P}_k)$
\EndFor
\end{algorithmic}
\end{algorithm}

% ============================================
% 5. EXPERIMENTAL SETUP
% ============================================
\section{Experimental Setup}
\label{sec:setup}

\subsection{Environment and Applications}

We evaluate across multiple desktop applications with varying MCP tool complexity (Table~\ref{tab:apps}).

\begin{table}[t]
\centering
\caption{Application benchmark overview. Each application has a distinct set of MCP tools, requiring the agent to adapt its modality selection strategy.}
\label{tab:apps}
\begin{tabular}{lcc}
\toprule
\textbf{Application} & \textbf{Seed Tasks} & \textbf{MCP Tools} \\
\midrule
VS Code & 30 & 9 \\
LibreOffice Calc & 45 & 16 \\
Chrome & 45 & 11 \\
\midrule
\textbf{Total} & \textbf{120} & \textbf{36} \\
\bottomrule
\end{tabular}
\end{table}

The applications span diverse domains: code editing (VS Code, 30 tasks), spreadsheet processing (LibreOffice Calc, 45 tasks), and web browsing (Chrome, 45 tasks). Task counts reflect the available evaluation benchmarks for each domain. MCP tool counts vary across applications, requiring the agent to learn application-specific modality strategies.

\subsection{Models and Training}

\paragraph{Base Model.}
We use \textbf{Qwen3-VL-8B} as our primary model, which supports native \texttt{<tool\_call>} format with high format accuracy. Cross-model analysis with additional base models is provided in Appendix~\ref{app:cross_model}.

\paragraph{Training and Evolution.}
We use LoRA \cite{hu2022lora} for parameter-efficient fine-tuning (rank 8, LR $2 \times 10^{-5}$). The evolution loop runs $K \in \{1, 3\}$ iterations with $N=3$ attempts per task and success threshold $\tau_{\text{success}} = 0.5$. Full training configuration details are provided in Appendix~\ref{app:sft_details}.

\subsection{Evaluation Protocol}

\paragraph{LLM-Judge Evaluation.}
Claude evaluates each trajectory, producing an LLM Score in $[0,1]$ that reflects task completion quality.

\paragraph{Metrics.}
We report \textbf{Pass Rate} (binary success from the LLM judge, determined independently of the continuous score), \textbf{LLM Score} (trajectory quality in $[0,1]$), and \textbf{MCP Ratio} (MCP vs.\ total actions).

% ============================================
% 6. RESULTS
% ============================================
\section{Results}
\label{sec:results}

We evaluate our framework across three desktop applications (Chrome, VS Code, LibreOffice Calc) with 120 tasks total (Table~\ref{tab:apps}), using Qwen3-VL-8B as the student model and Claude as the LLM judge.

\subsection{Single-Iteration Results on Chrome}
\label{sec:chrome_results}

Chrome represents an ideal setting for studying MCP--GUI synergy: tasks frequently map to direct MCP tool calls (e.g., \texttt{bookmark\_page}, \texttt{navigate\_to}), providing clean signal for tool invocation learning. Table~\ref{tab:chrome_main} compares policy learning strategies in a single iteration.

\begin{table}[t]
\centering
\caption{Chrome results after one iteration (Qwen3-VL-8B, 45 tasks). Multi-iteration trends in Table~\ref{tab:evolution}.}
\label{tab:chrome_main}
\small
\begin{tabular}{lcccc}
\toprule
\textbf{Strategy} & \textbf{Pass} & \textbf{Score} & \textbf{Eff.} & \textbf{MCP\%} \\
\midrule
Baseline & 60.0\% & 0.654 & 0.662 & 43.7\% \\
\midrule
\multicolumn{5}{l}{\textit{Experience Bank (inference-time only)}} \\
\quad + Exp. Bank & 62.2\% & 0.640 & 0.646 & 45.0\% \\
\midrule
\multicolumn{5}{l}{\textit{Trajectory Distillation}} \\
\quad + Distill. + Exp. & \textbf{77.8\%} & \textbf{0.770} & \textbf{0.724} & 36.3\% \\
\bottomrule
\end{tabular}
\end{table}

\textbf{Trajectory distillation yields substantial gains across all metrics.} Distillation achieves 77.8\% pass rate (+17.8pp over baseline), with 100\% on easy tasks (vs.\ 80\%) and 53.3\% on hard tasks (vs.\ 26.7\%). The model also achieves the highest LLM Score (0.770) and efficiency (0.724), learning more effective hybrid strategies that use fewer but better-targeted tool calls. \textbf{Experience bank provides complementary training-free gains.} Without any fine-tuning, experience augmentation achieves 62.2\% (+2.2pp), demonstrating that even lightweight inference-time intervention can improve task success.

\subsection{Self-Evolution Across Iterations}
\label{sec:evolution_results}

% A key contribution of our framework is the iterative self-evolution loop: each iteration refines the experience bank from accumulated trajectories and adds successful student trajectories to the training pool via rejection sampling. The pipeline also performs gap analysis to generate targeted tasks when weaknesses are identified. Table~\ref{tab:evolution} presents iteration-by-iteration results on Chrome (45 tasks).

A key contribution of our framework is the iterative self-evolution loop: at each iteration, the agent is first evaluated to produce a performance profile $\mathcal{P}_k$ (Section~\ref{sec:profiling}), which then drives gap analysis to automatically generate new targeted tasks $\mathcal{G}_k$ addressing identified weaknesses. Expert trajectories are then collected on these new tasks, the experience bank is refined, and successful student trajectories are added to the training pool via rejection sampling. Table~\ref{tab:evolution} presents iteration-by-iteration results on Chrome.

\begin{table}[t]
\centering
\caption{Self-evolution results across 3 iterations on Chrome (Qwen3-VL-8B, 45 tasks). Pass rate (\%) and LLM Score shown per iteration. Distillation achieves 77.8\% in a single iteration; the experience-only variant improves steadily to 64.4\% without fine-tuning.}
\label{tab:evolution}
\small
\begin{tabular}{lcccc}
\toprule
& \multicolumn{2}{c}{\textbf{Iter 0}} & \multicolumn{2}{c}{\textbf{Best Iter}} \\
\cmidrule(lr){2-3} \cmidrule(lr){4-5}
\textbf{Strategy} & \textbf{Pass} & \textbf{Score} & \textbf{Pass} & \textbf{Score} \\
\midrule
Exp. Bank Only & 60.0 & 0.654 & 64.4 (iter 2) & 0.674 \\
Distill. + Exp. & 60.0 & 0.654 & \textbf{77.8} (iter 1) & \textbf{0.770} \\
\bottomrule
\end{tabular}
\end{table}

\paragraph{Both variants improve through complementary mechanisms.}
The experience-only variant improves steadily to 64.4\% without fine-tuning, while distillation achieves a stronger peak (77.8\% at iter~1). Each mechanism targets different failure modes: distillation teaches correct tool invocation format and syntax, while the experience bank provides strategic shortcuts such as keyboard alternatives to unreliable GUI sequences. See Appendix~\ref{app:calc_evolution} (Calc) for detailed results.

\subsection{Cross-Application Analysis}
\label{sec:cross_app}

Table~\ref{tab:cross_app} compares strategies across applications with varying MCP--GUI composition, revealing the central insight of our work.

\begin{table}[t]
\centering
\caption{Cross-application comparison (pass rate \%). The optimal strategy depends on task characteristics: distillation excels for Chrome (+17.8pp), while the experience bank is more effective for GUI-intensive VS Code (+10.0pp) and Calc (+2.3pp). Task counts: VS Code 30, Chrome/Calc 45.}
\label{tab:cross_app}
\small
\begin{tabular}{lccc}
\toprule
& \textbf{Chrome (45)} & \textbf{VS Code (30)} & \textbf{Calc (45)} \\
\midrule
Baseline & 60.0 & 53.3 & 44.4 \\
+ Exp. Bank & 62.2 & \textbf{63.3} \gain{+10.0} & \textbf{46.7} \gain{+2.3} \\
+ Distill. + Exp. & \textbf{77.8} \gain{+17.8} & 43.3 & 42.2 \\
\bottomrule
\end{tabular}
\end{table}

\paragraph{Optimal strategy depends on task characteristics.}
The results validate our central claim (Contribution~1): the optimal evolution mechanism is determined by each application's task characteristics. For Chrome, where many tasks map to direct MCP tool calls, distillation teaches tool invocation patterns effectively (+17.8pp). For GUI-intensive VS Code, the experience bank's keyboard shortcut rules bypass unreliable visual grounding (+10.0pp). LibreOffice Calc, despite high baseline MCP usage (61.5\%), benefits modestly from the experience bank (+2.3pp)---its complex multi-step API chains require richer training signal than distillation from limited demonstrations provides.

\paragraph{Why distillation hurts certain applications.}
On VS Code, distillation \textit{decreases} pass rate by $-$10.0pp; on Calc, MCP usage drops from 61.5\% to 47.8\% after SFT. We observe evidence of \textbf{distribution mismatch}: on Calc, the student's baseline already achieves high MCP usage (61.5\%), but post-distillation MCP usage drops substantially, suggesting the expert trajectories shift the student toward a less MCP-heavy strategy. In contrast, the experience bank encodes \textit{abstract strategies} (e.g., ``use \texttt{Ctrl+Shift+P}'') that augment rather than override the model's existing policy. This suggests that when expert and student behavioral distributions diverge, inference-time augmentation may outperform imitation.

\paragraph{Experience bank provides consistent training-free improvement.}
Across all three applications, the experience bank improves over the baseline without any fine-tuning: +10.0pp on VS Code, +2.3pp on Calc, and +2.2pp on Chrome---demonstrating that \textit{inference-time experience augmentation} is a broadly applicable mechanism that complements rather than replaces trajectory distillation.

% ============================================
% 7. ANALYSIS
% ============================================
\section{Analysis}
\label{sec:analysis}

\subsection{Complementarity of Evolution Mechanisms}

The two evolution mechanisms address \textbf{complementary failure modes}, validating our hybrid policy formulation (Contribution~1): distillation teaches \textit{how} to invoke MCP tools (format, syntax), while the experience bank teaches \textit{when} to use each modality. Per-iteration profiling confirms this: distillation drives hard-task gains (+26.6pp) with MCP adoption rising from 43.7\% to 55.8\%, while the experience bank drives medium-task gains (+13.4pp; per-difficulty breakdowns in Appendix~\ref{app:profile}). The training pool grows from 44 to 79 samples via rejection sampling and gap-driven task generation, demonstrating the self-evolving loop's ability to autonomously grow training data. Full profiles are in Appendix~\ref{app:profile}.

\subsection{Qualitative Improvement Example}

We illustrate how the experience bank enables inference-time improvement (Contribution~3). On a Chrome print dialog task, the baseline agent called an incorrect MCP tool (\texttt{google\_chrome.print} with dot syntax), then spent 17 steps clicking randomly before failing. After one evolution iteration, the task was solved in a single step: the agent pressed \texttt{Ctrl+P}---a shortcut auto-discovered by the experience bank via trajectory comparison (Figure~\ref{fig:evolution_profile}, Iter~0). Before/after screenshots are in Appendix~\ref{app:qualitative}.

\subsection{What Does Self-Evolution Learn?}

\begin{figure}[t]
\centering
\small
\resizebox{\columnwidth}{!}{%
\begin{tikzpicture}[
    node distance=0.3cm,
    profilebox/.style={rectangle, draw, rounded corners=4pt, minimum width=2.7cm, align=left, font=\scriptsize, inner sep=3pt},
    mechbox/.style={rectangle, draw, rounded corners=3pt, fill=evolvecolor!8, minimum width=3.3cm, align=left, font=\scriptsize, inner sep=3pt},
    deltabox/.style={rectangle, draw=evolvecolor, line width=0.8pt, rounded corners=3pt, fill=evolvecolor!5, minimum width=2.7cm, align=left, font=\scriptsize, inner sep=3pt},
    arrow/.style={->, >=stealth, very thick, color=evolvecolor},
]

% LEFT: Baseline Profile
\node[profilebox, fill=gray!8] (baseline) {
\textbf{Baseline Profile}\\[2pt]
Pass rate: 60.0\%\\
E 80 / M 73 / H 27\%\\
MCP ratio: 43.7\%\\
Total actions: 243\\[2pt]
\textit{Weaknesses:}\\
\textbullet~GUI-heavy exploration\\
\textbullet~Wrong MCP syntax\\
~~~~(\texttt{chrome.print} fails)\\
\textbullet~0/15 hard tasks via MCP
};

% CENTER: Evolution Mechanisms
\node[mechbox, right=0.25cm of baseline] (evolution) {
\textbf{Self-Evolution Loop}\\[2pt]
\textit{Gap Analysis} $\to$ \textit{Targeted Tasks}\\
\textbullet~Underused tools identified:\\
~~~~\texttt{bookmark\_page},\\
~~~~\texttt{delete\_browsing\_data}\\
\textbullet~New tasks generated\\[2pt]
\textit{Experience Bank (6 categories)}\\
\textbullet~\texttt{Ctrl+P} for print dialog\\
\textbullet~MCP uses \texttt{\_\_} not dots\\
\textbullet~\texttt{Ctrl+Shift+T} restore tab\\
\textbullet~After \texttt{open\_privacy},\\
~~~~continue with GUI steps\\[2pt]
\textit{Training Pool}\\
\textbullet~44 $\to$ 79 samples\\
\textbullet~Rejection sampling +\\
~~~~gap-driven generation
};

% RIGHT: Improved Profile
\node[deltabox, right=0.25cm of evolution] (improved) {
\textbf{Best Achieved}\\[2pt]
\textit{Peak pass rate (iter 1):}\\
Pass: \textbf{77.8\%}~~{\color{evolvecolor}$\uparrow$17.8}\\
E \textbf{100} / M 80 / H \textbf{53\%}\\[2pt]
\textit{Efficiency gains (iter 3):}\\
MCP ratio: \textbf{55.8\%}~~{\color{evolvecolor}$\uparrow$12.1}\\
Actions: \textbf{163}~~{\color{evolvecolor}$\downarrow$33\%}\\
M \textbf{87\%}~~{\color{evolvecolor}$\uparrow$13.4}\\[2pt]
\textit{Key improvements:}\\
\textbullet~Hard tasks: 27$\to$53\%\\
\textbullet~MCP-first strategies\\
~~~~replace GUI exploration
};

% Arrows
\draw[arrow] (baseline.east) -- (evolution.west);
\draw[arrow] (evolution.east) -- (improved.west);

\end{tikzpicture}%
}%
\caption{Self-evolution improvement on Chrome (Qwen3-VL-8B, Distillation + Experience Bank). \textbf{Left}: Baseline profile with identified weaknesses. \textbf{Center}: Evolution mechanisms---gap analysis, experience bank, and training pool growth. \textbf{Right}: Best metrics achieved, separated by source iteration: peak pass rate at iter~1 (+17.8pp), efficiency gains by iter~3 (MCP +12.1pp, actions $-$33\%). Full profile comparison in Appendix~\ref{app:profile}.}
\label{fig:evolution_profile}
\end{figure}
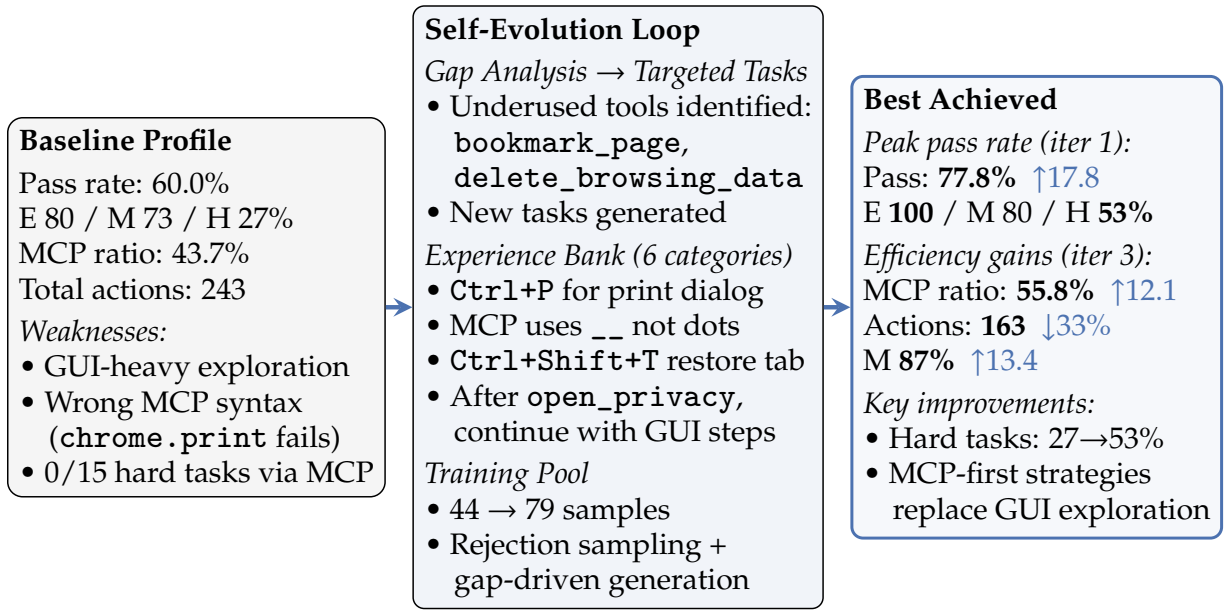

Figure~\ref{fig:evolution_profile} summarizes the end-to-end improvement achieved by the self-evolution pipeline (Contribution~2). Starting from a baseline that relies on GUI exploration with incorrect MCP syntax, the pipeline autonomously identifies weaknesses, generates targeted tasks, and accumulates experience---all without manual intervention. All experience rules are auto-discovered by LLM trajectory comparison---none are manually programmed.

Three key improvements emerge.
First, \textbf{MCP adoption increases} from 43.7\% at baseline to 55.8\% by iteration~3, indicating the agent progressively learns to prefer structured API calls over GUI alternatives (note: iteration~1 shows a temporary dip to 36.3\% as the model first learns tool syntax before optimizing tool selection).
Second, \textbf{total actions decrease by 33\%} (243$\to$163), reflecting more efficient execution as the agent avoids exploratory GUI sequences.
Third, the \textbf{gap analysis adapts}: early iterations identify format errors (dot vs.\ underscore syntax), while later iterations target underused tools (e.g., bookmark management), showing that profiling surfaces progressively finer-grained weaknesses.

\subsection{Cross-Model Comparison}

We compare Qwen3-VL-8B and Qwen2.5-VL-7B on Chrome to identify prerequisites for effective self-evolution. Qwen2.5 achieves 48.9\% baseline but \textit{degrades} with experience augmentation ($-$8.9pp), while Qwen3 improves (+2.2pp experience, +17.8pp distillation). The key differentiator is native tool-calling support: Qwen3's \texttt{<tool\_call>} format enables MCP invocations, while Qwen2.5 treats structured tool syntax as noise (6.6\% vs.\ 43.7\% MCP ratio). The agent must have base capability in \textit{both} modalities for self-evolution to amplify performance. Detailed analysis is in Appendix~\ref{app:cross_model}.

% ============================================
% 8. DISCUSSION
% ============================================
\section{Discussion}
\label{sec:discussion}

\paragraph{Application-Aware Mechanism Selection.}
Our cross-application analysis reveals that the same training mechanism can help or hinder depending on task characteristics. Distillation excels when the student can reproduce expert tool calls but degrades for GUI-intensive tasks where visual grounding varies across environments. Practitioners should thus select mechanisms based on their target application's MCP--GUI composition.

\paragraph{Data Efficiency and Experience Bank Design.}
The pipeline bootstraps from only 44 expert demonstrations, autonomously growing to 79 training samples via rejection sampling and gap-driven generation---substantially fewer environment interactions than online RL approaches~\cite{bai2024digirl}. Two design decisions proved critical: \textbf{per-category capacity limits} prevent context overflow in smaller models, and \textbf{application-type filtering} prevents cross-app contamination.

\paragraph{Limitations and Future Work.}
The framework requires an expert model for trajectory seeding and assumes LLM-judge reliability for quality filtering. The experience bank's effectiveness depends on instruction-following capability, as shown by Qwen2.5's inability to leverage augmented context. LibreOffice Calc reveals a compositional complexity boundary where multi-step API chains require richer training signal than single-step tool calls. Promising directions include integrating online exploration with reward signals from the LLM judge and improving distillation for GUI-intensive tasks.

% ============================================
% 9. CONCLUSION
% ============================================
\section{Conclusion}
\label{sec:conclusion}

We presented a self-evolving framework for hybrid MCP-GUI agents combining distillation (+17.8pp on Chrome), inference-time experience augmentation (+10.0pp VS Code), and automatic environment generation. The framework iteratively identifies weaknesses and generates targeted tasks, establishing a practical paradigm for self-improving computer-use agents.

% ============================================
% REFERENCES
% ============================================

\bibliographystyle{splncs04}
\bibliography{main}

% ============================================
% APPENDIX
% ============================================
\appendix

\setcounter{section}{0}
\renewcommand{\thesection}{\Alph{section}}

\section{Skill Category Definitions}
\label{app:skills}

Table~\ref{tab:skills} provides detailed skill category definitions with cross-application examples. These six application-agnostic categories form the basis of our multi-dimensional performance profiling and experience bank organization.

\begin{table}[H]
\centering
\small
\caption{Application-agnostic skill categories with examples.}
\label{tab:skills}
\begin{tabular}{p{3cm}p{9cm}}
\toprule
\textbf{Skill} & \textbf{Example Tasks} \\
\midrule
Data Retrieval & Read file (VS Code), Get cell (Calc), Get page (Chrome) \\
Data Manip. & Write code (VS Code), Edit cell (Calc), Fill form (Chrome) \\
Search  & Find text (VS Code), Search cells (Calc), Search page (Chrome) \\
Execution & Run tests (VS Code), Calculate (Calc), Navigate URL (Chrome) \\
Navigation & Go to file (VS Code), Switch sheet (Calc), Switch tab (Chrome) \\
Configuration & Change theme (VS Code), Set format (Calc), Preferences (Chrome) \\
\bottomrule
\end{tabular}
\end{table}

\section{Implementation Details}
\label{app:implementation}

\subsection{Hyperparameters}
\label{app:sft_details}

Table~\ref{tab:hyperparams} lists all pipeline and SFT hyperparameters used in our experiments.

\begin{table}[h]
\centering
\caption{Pipeline and SFT hyperparameters.}
\small
\begin{tabular}{lrl}
\toprule
\textbf{Parameter} & \textbf{Value} & \textbf{Description} \\
\midrule
Attempts/task & 3 & Max attempts \\
Success threshold & 0.5 & Action success rate \\
Learning rate & 2e-5 & SFT learning rate \\
LoRA rank & 8 & LoRA rank \\
Image pixels & 50,176 & Per-image resolution \\
Max images & 30 & Per training sample \\
Cutoff length & 32,768 & Max sequence length \\
\bottomrule
\end{tabular}
\label{tab:hyperparams}
\end{table}

\subsection{Evolution Memory Structure}

The evolution memory module tracks:
\begin{itemize}
    \item \textbf{TaskMemory}: Per-task success/failure across iterations
    \item \textbf{IterationMemory}: Per-iteration summary statistics
    \item \textbf{EvolutionMemory}: Global patterns and experience extraction
\end{itemize}

\section{LLM Judge Prompt}
\label{app:judge_prompt}

The LLM judge (Claude) evaluates each trajectory using the following prompt template:

\begin{verbatim}
You are evaluating a computer use agent's performance.

Task: {task_description}
Application: {application_name}
Available MCP Tools: {tool_list}
Agent Trajectory: {trajectory}

Evaluate:
1. Task Completion (0-1): Did the agent achieve the goal?
2. Action Correctness: Were actions appropriate?
3. Tool Usage: Did the agent use MCP tools effectively?
4. Efficiency: Steps vs optimal?

Provide assessment as JSON:
{
  "score": 0.0-1.0,
  "success": true/false,
  "mcp_actions": <count>,
  "gui_actions": <count>,
  "reasoning": "..."
}
\end{verbatim}

\section{Application Environments}
\label{app:environments}

Our evaluation spans three desktop applications with distinct MCP--GUI interaction patterns:

\begin{itemize}
\item \textbf{VS Code}: 30 tasks across code editing scenarios, a GUI-intensive environment where tasks require precise visual interaction. Tasks include file navigation, code editing, terminal usage, and configuration changes.
\item \textbf{LibreOffice Calc}: 45 tasks across spreadsheet processing, with the highest baseline MCP usage (61.5\%). Tasks involve cell manipulation, formula computation, formatting, and data analysis.
\item \textbf{Chrome}: 45 tasks across web browsing scenarios, where many tasks map to direct MCP tool calls (baseline MCP ratio: 43.7\%). Tasks include navigation, bookmark management, form filling, and privacy settings.
\end{itemize}

Each application has a distinct MCP--GUI composition requiring application-aware strategy selection.

\section{Detailed Chrome Results}
\label{app:chrome_detailed}

Table~\ref{tab:chrome_by_difficulty} breaks down Chrome results by task difficulty, showing that distillation provides the largest gains on hard tasks.

\begin{table}[h]
\centering
\caption{Chrome results by difficulty level (Qwen3-VL-8B, 45 tasks: 15 easy, 15 medium, 15 hard).}
\label{tab:chrome_by_difficulty}
\small
\begin{tabular}{lcccc}
\toprule
\textbf{Strategy} & \textbf{Easy} & \textbf{Med.} & \textbf{Hard} & \textbf{Score} \\
\midrule
Baseline & 12/15 & 11/15 & 4/15 & 0.654 \\
+ Exp. Bank & 14/15 & 10/15 & 4/15 & 0.640 \\
+ Distill. + Exp. & \textbf{15/15} & \textbf{12/15} & \textbf{8/15} & \textbf{0.770} \\
\bottomrule
\end{tabular}
\end{table}

Distillation achieves 100\% on easy tasks (vs.\ 80\% baseline) and doubles hard task performance (53.3\% vs.\ 26.7\%), demonstrating that MCP tool invocation patterns transfer effectively through trajectory distillation.

\section{LibreOffice Calc Evolution Results}
\label{app:calc_evolution}

Table~\ref{tab:calc_evolution} compares baseline and best-achieved results on LibreOffice Calc. Unlike Chrome, Calc shows limited improvement from self-evolution despite high baseline MCP usage (61.5\%).

\begin{table}[h]
\centering
\caption{Calc self-evolution results (Qwen3-VL-8B, 45 tasks). The experience bank provides a marginal gain (+2.3pp); distillation does not improve over baseline. Hard tasks (0/15) remain unsolved across all strategies.}
\label{tab:calc_evolution}
\small
\begin{tabular}{lcc}
\toprule
\textbf{Strategy} & \textbf{Baseline} & \textbf{Best} \\
\midrule
Exp. Bank & 44.4\% & \textbf{46.7\%} \\
Distill. + Exp. & 44.4\% & 44.4\% \\
\bottomrule
\end{tabular}
\end{table}

We attribute Calc's resistance to evolution to the complexity of its MCP tool sequences, which require multi-step API chains (reading cell ranges, computing formulas, writing results). These compositional tool patterns are harder to learn from limited demonstrations than Chrome's simpler, more direct tool invocations.

\section{Experience Bank Design Analysis}
\label{app:exp_design}

Our experiments reveal five design principles for experience banks targeting small (7--8B) language models:

\textbf{(1) Conciseness over coverage.} Category-specific experience banks must be capped to prevent context overflow. In our ablations, uncapped banks degraded performance because verbose experience text competed with task-relevant information for the model's attention window. The LLM-based merge and summarization process ensures only the most transferable patterns are retained.

\textbf{(2) Actionable specificity.} Rules must prescribe \textit{concrete actions} rather than abstract strategies. For example, ``Use Ctrl+P to open the print dialog'' directly improved performance on the print task, while abstract advice like ``be efficient with tool usage'' had no measurable effect. The LLM trajectory comparison naturally produces specific rules because it identifies the exact action differences between successful and failed trajectories.

\textbf{(3) Domain isolation.} Banks must be filtered by application type to prevent cross-app contamination. In early experiments without filtering, VS Code keyboard shortcuts (e.g., ``Ctrl+Shift+E for Explorer'') appeared in Chrome experience banks, causing the agent to execute irrelevant shortcuts. Adding the application-type filter to the experience bank builder eliminated this issue.

\textbf{(4) Distillation and experience bank are complementary.} Distillation provides rapid single-iteration gains (77.8\% at iter~1) while the experience bank enables steady training-free improvement (60.0\%$\to$64.4\%). This suggests the two mechanisms address orthogonal failure modes---distillation corrects tool invocation errors while the experience bank corrects planning and strategy errors.

\textbf{(5) Automatic rule discovery over manual engineering.} All rules in the experience bank were discovered automatically by the LLM comparing trajectory pairs. Manual rule engineering does not scale across applications and cannot capture the nuanced interaction patterns (e.g., ``after calling \texttt{open\_privacy\_settings}, continue with GUI steps'') that emerge from real agent-environment interactions.

\section{Qualitative Before/After Example}
\label{app:qualitative}

We provide a concrete example of behavioral improvement through self-evolution on the Chrome print dialog task (\texttt{chrome\_easy\_07}: ``Open the print dialog for this page'').

\textbf{Before evolution (Iter 0).} The baseline agent attempts to print via the File menu, triggering a ``Save File'' dialog instead of the print dialog. It then spends 17 steps trying to navigate the file save interface, never reaching the correct print dialog.

\textbf{After evolution (Iter 1+).} After the experience bank learns ``Use Ctrl+P to open print dialog'', the agent executes the keyboard shortcut directly, reaching the correct Chrome print dialog in a single step.

This example illustrates the core mechanism: the experience bank distills \textit{actionable patterns} from trajectory comparison that directly improve task execution. The keyboard shortcut Ctrl+P was never explicitly programmed---it was discovered by the LLM comparing successful and failed trajectories, then stored as a transferable rule in the Navigation skill category.

\section{Pipeline Details}
\label{app:pipeline_details}

The self-evolution pipeline orchestrates all components automatically. It handles environment setup with health validation, automatic failure recovery, dynamic port allocation for parallel GPU execution, MCP server lifecycle management across Docker containers, and application-specific workspace reset procedures.

\section{Cross-Model Analysis}
\label{app:cross_model}

Table~\ref{tab:model_compare} compares Qwen3-VL-8B and Qwen2.5-VL-7B on Chrome (45 tasks), demonstrating that self-evolution amplifies existing capabilities but cannot create new ones.

\begin{table}[h]
\centering
\caption{Cross-model comparison on Chrome (45 tasks). Self-evolution mechanisms only improve Qwen3-VL-8B; Qwen2.5-VL-7B shows no improvement from any intervention.}
\label{tab:model_compare}
\small
\begin{tabular}{lcc}
\toprule
\textbf{Strategy} & \textbf{Qwen3-8B} & \textbf{Qwen2.5-7B} \\
\midrule
Baseline & 60.0\% & 48.9\% \\
+ Exp. Bank & 62.2\% & 40.0\% \\
+ Distill. + Exp. & \textbf{77.8\%} & --- \\
\midrule
Native \texttt{<tool\_call>} & \ding{51} & \ding{55} \\
Instruction following & Strong & Weak \\
\bottomrule
\end{tabular}
\end{table}

Qwen2.5-VL-7B lacks native tool-calling support and shows systematic capability gaps: minimal MCP usage (6.6\% MCP ratio vs.\ 43.7\% for Qwen3), no reasoning before actions, and early termination. Experience augmentation actually degrades performance ($-$8.9pp), as the model treats experience text as noise. Distillation is not applicable to Qwen2.5 because it lacks the native \texttt{<tool\_call>} format required for SFT with tool-calling trajectories. This demonstrates that self-evolution requires a minimum level of base model capability---particularly native tool-calling support---to be effective.

\section{Additional Analysis}
\label{app:additional_analysis}

\subsection{Gap-Generated Task Examples}

The gap analysis identified that the model was GUI-heavy (low MCP utilization) and underused tools like \texttt{bookmark\_page}, \texttt{bring\_back\_last\_tab}, and \texttt{delete\_browsing\_data}. Generated tasks target these specific gaps with easy difficulty to build foundational MCP tool usage.

\subsection{Training Pool Growth}

Profile-guided task generation identifies systematic weaknesses across modality, difficulty, and skill dimensions. At iteration~2, gap analysis identified the agent as GUI-heavy with underused MCP tools, generating targeted tasks with expert demonstrations collected for each. On Chrome, the training pool grows from 44 initial expert demonstrations to 79 samples by iteration~3 through a combination of rejection sampling (which adds successful student trajectories) and gap-driven generation. This self-amplification diversifies the training signal beyond the initial expert seed set.

\section{Evolution Profile Analysis}
\label{app:profile}

A key advantage of multi-dimensional performance profiling is that it reveals \textit{how} the agent improves across iterations, not just whether it improves. Distillation + Experience Bank provides the largest gains on pass rate (+17.8pp) and hard tasks (+26.6pp), while Experience Bank Only delivers steady improvement on medium tasks (+13.4pp) without fine-tuning.

\textbf{(1) MCP adoption as a learning signal.} Distillation increases MCP tool usage from 43.7\% to 55.8\%, accompanied by a 33\% reduction in total actions (243$\to$163). This indicates the model learns to replace lengthy GUI exploration with targeted API calls---the core benefit of trajectory distillation.

\textbf{(2) Difficulty-specific improvement patterns.} Distillation provides the largest gains on hard tasks (+26.6pp), while the experience bank primarily improves medium tasks (+13.4pp). Easy tasks saturate quickly under both strategies ($\geq$80\% from baseline). This suggests that hard tasks require learning new tool invocation patterns (distillation), while medium tasks benefit from strategic guidance (experience bank).

\textbf{(3) Training pool growth.} The distillation experiment's training pool grows from the initial 44 expert demonstrations to 79 samples through rejection sampling and gap-driven task generation, diversifying the training signal beyond the initial expert seed set.

\end{document}